\def\year{2018}\relax
\pdfoutput=1
\documentclass[letterpaper]{article} 
\usepackage{aaai18}  
\usepackage{times}  
\usepackage{helvet}  
\usepackage{courier}  
\usepackage{url}  
\usepackage{graphicx}  
\usepackage{float}
\usepackage{algorithm}
\usepackage{algorithmic}
\usepackage{times}

\usepackage{amsmath}
\usepackage{amssymb}
\usepackage{bm}
\usepackage{caption}

\usepackage{epsfig}
\usepackage{epstopdf}
\usepackage{subfigure}
\usepackage{array}
\usepackage{multirow}
\usepackage{eucal}
\usepackage{makecell}
\usepackage{fnbreak} 

\newcommand{\xucomment}[1]{}

\begin{document}
%
\title{BT-Nets: Simplifying Deep Neural Networks via Block Term Decomposition}
\author{
Guangxi Li$^{1}$, Jinmian Ye$^1$, Haiqin Yang$^2$, Di Chen$^1$, Shuicheng Yan$^3$, Zenglin Xu$^{1,*}$\\
$^1$SMILE Lab, School of Comp. Sci. and Eng., Univ. of Elec. Sci. and Tech. of China\\
$^2$Department of Computing, Hang Seng Management College\\
$^3$National University of Singapore, 360 AI Institute\\
$^*$Corresponding author: zenglin@gmail.com
}
\maketitle
\begin{abstract}

Recently,  deep neural networks (DNNs) have been regarded as the state-of-the-art classification methods in a wide range of applications, especially in image classification. Despite the success, the huge number of parameters blocks its deployment to situations with light computing resources. Researchers resort to the redundancy in the weights of DNNs and attempt to find how fewer parameters can be chosen while preserving the accuracy at the same time. 
Although several promising results have been shown along this research line, most existing methods either fail to significantly compress a well-trained deep network or require a heavy fine-tuning process for the compressed network to regain the original performance. In this paper, we propose the \textit{Block Term} networks (BT-nets) in which the commonly used fully-connected layers (FC-layers) are replaced with block term layers (BT-layers). In BT-layers, the inputs and the outputs are reshaped into two low-dimensional high-order tensors, then  block-term decomposition is applied as tensor operators to connect them. We conduct extensive experiments on benchmark datasets to demonstrate that BT-layers can achieve a very large compression ratio on the number of parameters while preserving the representation power of the original FC-layers as much as possible. Specifically, we can get a higher performance while requiring fewer parameters compared with the tensor train method. 

\end{abstract}

\section{Introduction}



Deep neural networks (DNNs) have achieved  significantly improved performance in a wide range of applications, such as image classification, speech recognition, natural language processing, etc. Specifically, the most famous DNN architectures in image classification include: AlexNet \cite{krizhevsky2012imagenet}, VGGNet \cite{simonyan2014very}, and so on.
They all won the championship on the ImageNet dataset \cite{deng2012imagenet} of that year.
However, due to their complex structure, multiple layers and the huge amount of parameters, DNNs  have extremely higher spatial and temporal complexity compared with classic machine learning models. As a result, it usually takes several days to train even on powerful Graphics Processing Units (GPU) and the trained model also takes a large memory. For example, AlexNet takes five to six days to train on two GTX 580 3GB GPUs and its Caffe model size is up to 240MB \cite{jia2014caffe}.
Therefore, compressing DNN architectures to decrease the temporal and spatial complexity is becoming an important issue that calls for urgent solutions.

Since it has been proved that there is huge redundancy in the weights of DNNs \cite{zeiler2014visualizing}, so it is possible to compress DNNs without or with little reduction in accuracy.
As the flow data with tensor structure is ubiquitous in DNNs and tensor decomposition methods (or tensor networks) have long been the subject of theoretical study, it is natural to investigate the intersection between them.

Recently, tensor decomposition methods are used to compress DNNs mainly in the following three lines. One is speeding up the convolutional layers of DNNs several times via low-rank tensor approximation methods at the cost of small accuracy drop, e.g., 1\% \cite{lebedev2014speeding,denton2014exploiting}. 
The second line is compressing the fully-connected layers (FC-layers), which have the largest amount of parameters (usually more than 80 percent of the total number of parameters in both popular networks: AlexNet and VGG), by replacing them with tensor layers \cite{novikov2015tensorizing,kossaifi2017tensor}.
The last line is attempting to connect the neural networks with the tensor networks. In \cite{cohen2016expressive}, an equivalence between some neural  networks and hierarchical tensor factorization is established.

Since compressing the part that possesses the largest amount of parameters tends to be most effective, we focus on compressing the FC-layers.
\cite{kossaifi2017tensor} replaces the FC-layers with \textit{Tensor Contraction} layers (TC-layers) where we could output a low dimensional projection of the input tensor. However, the compression ratio is not big enough due to the absence of reshaping operation on the input tensors. \cite{novikov2015tensorizing} similarly substitutes the \textit{Tensor Train} layers (TT-layers) for the FC-layers  using the matrix product operator where a huge compression factor of the number of parameters in FC-layers can be achieved. But the reduction in accuracy is rather large, especially when the TT-ranks are small, because the connections (TT-ranks) are quite weak.
So, what is the least amount of parameters that are necessary in the limit case?
Can we change the connection methods to get stronger connections with a fewer number of parameters and smaller accuracy drop at the same time? This paper tries to answer this question.

 We propose \textit{Block Term} layers (BT-layers), which are based on block term decomposition \cite{de2008decompositions}, to substitute the FC-layers for the purpose that the number of parameters in these FC-layers can be extremely reduced. 
 BT decomposition combines Tucker decomposition \cite{tucker1966some} and CANDECOMP/PARAFAC decomposition \cite{harshman1970foundations} and makes a trade-off to share the advantages of both. 
 We call a DNN that contains BT-layers BT-Net. The contributions of BT-Nets are concluded as follows:
\begin{itemize}
  \item BT-layers are able to preserve the representation power of FC-layers as much as possible thank to their \textit{exponential representation ability} trait. That is when more input modes are taken into consideration, we will automatically get the exponential growth of the BT-ranks. Therefore,  BT-layers have stronger connections.
  \item In BT-Nets, there is no need to fine-tune the BT-ranks because of the \textit{commutativity} of  BT-layers. By comparison, in TT-layers, TT-ranks usually have the \textit{olive} property, which means the values must be small at both ends and large in the middle. So we need to do fine-tuning process if we want to obtain the best performance.
  \item The experiments demonstrate that we can get a very large compression ratio in the number of parameters with tolerable impacts on the accuracy.
\end{itemize}

\section{Related Work}

There are several methods proposed to compress deep neural networks in these years. \cite{chen2015compressing}  compresses  deep neural networks by employing the hashing trick, because there is redundancy in the weight of DNN \cite{zeiler2014visualizing}. \cite{han2015deep} compresses deep neural networks with pruning, trained quantization and huffman coding, which achieves a high compression ratio in terms of storage.
\cite{hubara2016binarized} proposes the binarized neural networks where the parameters and the activation functions are binarized  in the training process and the gradients are binarized as well. This work can speedup 7 times compared with the traditional CNN.
In addition, there are also some methods aiming to speed up the convolutional layers via low-rank decomposition of the convolutional kernel tensor \cite{lebedev2014speeding,denton2014exploiting,jaderberg2014speeding,kim2015compression}.

Recently, tensor decomposition is getting more and more attention in a wide range of fields, such as signal processing \cite{de2007fourth}, computer vision \cite{shashua2001linear}, numerical analysis \cite{beylkin2002numerical},  and network analysis or recommendation systems \cite{tensorXuYQ15,chen2013exact,LiXWYKL17}. A comprehensive overview about the tensor decomposition can be found in the survey paper \cite{kolda2009tensor}. 
Tensor decomposition has recently been studied in the connection with deep learning, like sharing residual units \cite{yunpeng2017sharing}, speeding up CNNs \cite{lebedev2014speeding}, tensor switching nets \cite{tsai2016tensor}, tensorizing neural networks \cite{novikov2015tensorizing,garipov2016ultimate} and tensor regression networks \cite{kossaifi2017tensor1}.

Concurrent with our work, \cite{kossaifi2017tensor} proposes the TC-layers to substitute the FC-layers, which just perform a transformation along each mode of the input tensors. Therefore, the compression ratio of the parameters is not quite high.
\cite{novikov2015tensorizing} replaces the FC-layers with TT-layers, where the input tensors are reshaped into low-dimensional high-order tensors and the weight matrices are factorized into the corresponding tensor train format.  
Compared with these works, we use the block-term format (BT-format) to replace the weight matrices in the FC-layers, where we can get stronger connections (i.e., higher BT-ranks) with fewer parameters.

It is important to note that there are some neural network architectures without large FC-layers  such as the ResNet \cite{he2016deep}, which aims to designing narrower and deeper networks. The motivation of our proposed model is different from these networks and it is possible to combine the proposed BT-layers with the ResNet. 



\section{Block Term Format}

\subsection{Tensor Network Diagrams and Notations}

\begin{figure}[t]
  \centering
 \includegraphics[width=0.48\textwidth]{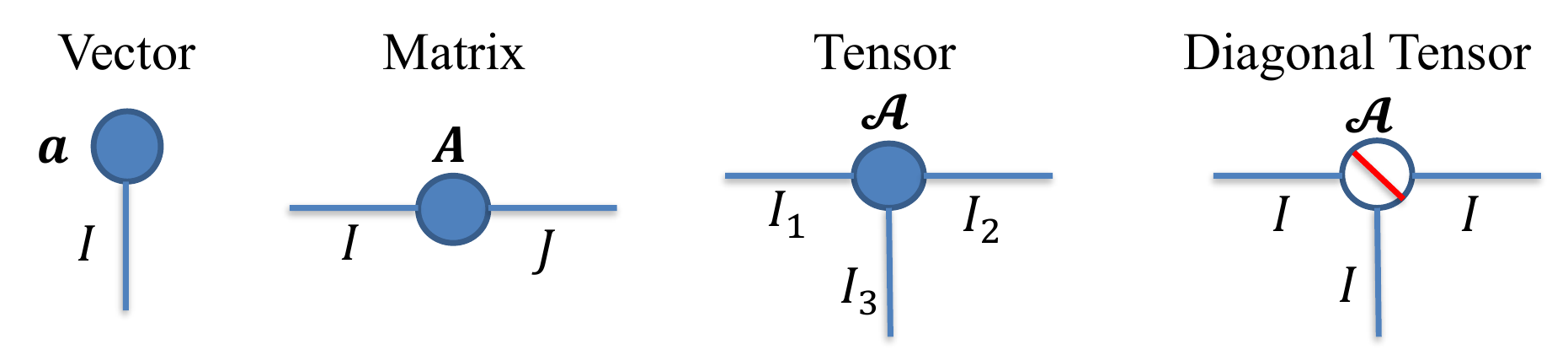} 
 \vspace{-2ex}
 \caption{Basic symbols for tensor network diagrams. } \label{fig:m:basic_symbol}
\end{figure}

\begin{figure*}[t]
\centering
\subfigure[Tensor Unfolding]{\label{fig:m:tensor_unfold}
 \includegraphics[width=0.4\textwidth]{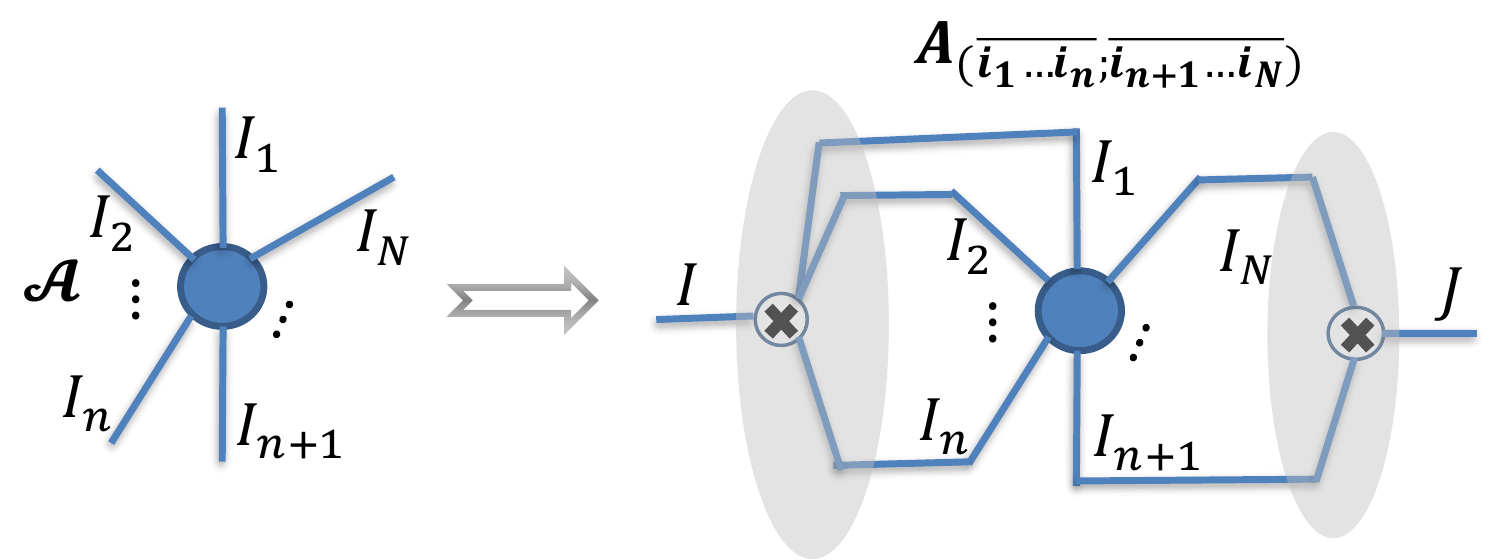}}
\subfigure[Tensor Contraction]{\label{fig:m:tensor_contract}
 \includegraphics[width=0.45\textwidth]{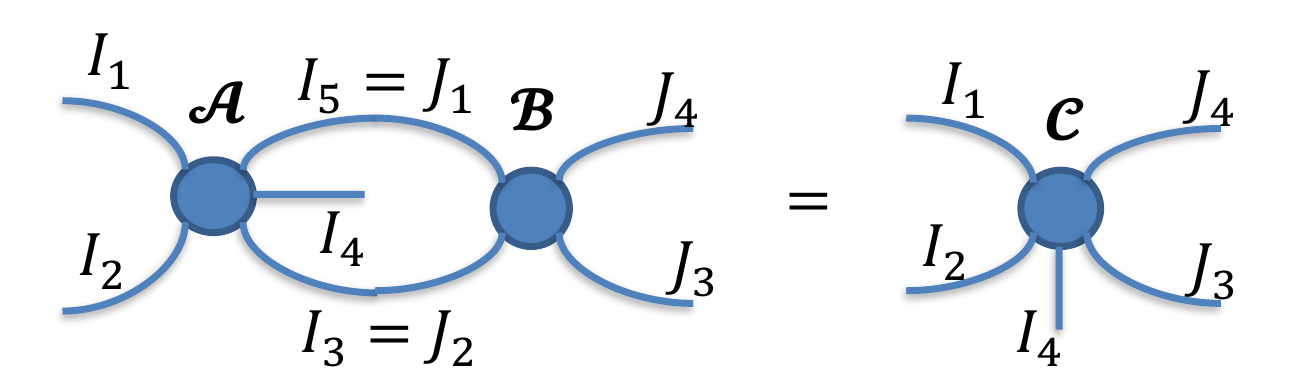}}
 \vspace{-2ex}
  \caption{Diagrams for tensor unfolding and tensor contractions.  (a) Given an order-$N$ tensor $\bm{\mathcal A}\in\mathbb R^{I_1\times I_2 \ldots\times I_N}$, its mode-([$n$]) unfolding yields a matrix $\bm A_{(\overline{i_1\ldots i_n};\overline{i_{n+1}\ldots i_N})}\in \mathbb R^{I_1I_2 \ldots I_n\times I_{n+1}\ldots I_N}$; (b) Given $\bm{\mathcal A}\in\mathbb R^{I_1\times I_2\times I_3\times I_4\times I_5}$ and $\bm{\mathcal B}\in\mathbb R^{J_1\times J_2\times J_3\times J_4}$, the tensor contraction between them yield $\bm{\mathcal C}\in\mathbb R^{I_1\times I_2\times I_4\times J_3\times J_4}$.} \label{fig:m:tensor_unfold_contract}
 \end{figure*}

A tensor in tensor network, also known as a multi-way array, can be viewed as a higher-order extension of a vector (i.e., an order-1 tensor) and a matrix (i.e., an order-2 tensor).  Like rows and columns in a matrix, an order-$N$ tensor has $N$ modes whose lengths are represented by $I_1$ to $I_N$, respectively. The basic operations of tensors include linear arithmetic, tensor product, tensor transpose and tensor contraction. When the amount of tensors is big and the contraction relationships among their indices are complicated, a better way to represent them is using diagrams, namely tensor network diagrams. The basic symbols for tensor network diagrams are shown in Fig.~\ref{fig:m:basic_symbol}, in which tensors are  denoted graphically by nodes and edges. Each edge emerged from a node denotes a mode (or order, index) \cite{cichocki2014era}. Here, we use boldface lowercase letters, e.g., $\bm a$, to denote vectors, boldface capital letters, e.g., $\bm A$, to denote matrices and boldface Euler script letters, e.g., $\bm{\mathcal A}$, to denote higher-order tensors (order-3 or higher), respectively. 

\subsubsection{Tensor Unfolding}
Tensor unfolding, also called matricization, is virtually flattening (or reshaping) a tensor into a large dimensional matrix. To be more specific, given an order-$N$ tensor $\bm{\mathcal A}\in\mathbb R^{I_1\times I_2 \ldots\times I_N}$, its mode-([$n$]) unfolding yields a matrix $\bm A_{(\overline{i_1\ldots i_n};\overline{i_{n+1}\ldots i_N})}\in \mathbb R^{I_1I_2 \ldots I_n\times I_{n+1}\ldots I_N}$ such that the indices in the two parts (rows and columns) are arranged in a specific order, e.g., in the lexicographical order \cite{kolda2009tensor}, respectively, see Fig.~\ref{fig:m:tensor_unfold}. However, there is no need to force the first $n$ indices of $\bm{ \mathcal A}$ to be set to the rows (the first part of the indices) of the unfolding, on the contrary, the rows can be set with an arbitrary $n$ different indices of $\bm{ \mathcal A}$. So, in a more general case, let $\bm r= \{m_1,m_2,\ldots,m_R\} \subset \{1,2,\dots,N\}$ be the row indices and $\bm c= \{n_1,n_2,\ldots,n_C\} \subset \{1,2,\dots,N\}-\bm r$ be the column indices, then the mode-($\bm r$, $\bm c$) unfolding of $\bm{ \mathcal A}$ produces a matrix  $\bm A_{(\bm r, \bm c)}\in \mathbb R^{I_{m_1}I_{m_2} \ldots I_{m_R}\times I_{n_1}I_{n_2}\ldots I_{n_C}}$.
\subsubsection{Tensor Contraction}
Tensor contraction between two tensors means that they are contracted into one tensor along the associated pairs of indices as illustrated in Fig.~\ref{fig:m:tensor_contract}. The whole process of tensor contraction between two tensors mainly contains the following three steps: i) perform tensor unfoldings upon them along the connected indices (for example, in Fig.~\ref{fig:m:tensor_contract}, we unfolds  $\bm{ \mathcal A}$ and  $\bm{ \mathcal B}$ as $\bm A_{(\overline{i_1i_2i_4};\overline{i_3i_5})}$ and $\bm B_{(\overline{j_2j_1};\overline{j_3j_4})}$, respectively), ii) perform a matrix multiplication between the two unfoldings, iii) perform a reshape operation on the matrix product.
Tensor contractions among multiple tensors (e.g., tensor networks) can be computed by performing tensor contraction between two tensors many times. Hence, the order (number of modes) of an entire tensor network is given by the number of dangling (free) edges that is not contracted.

\subsection{Block Term Decomposition}

\begin{figure*}[t]
\centering
 \includegraphics[width=0.98\textwidth]{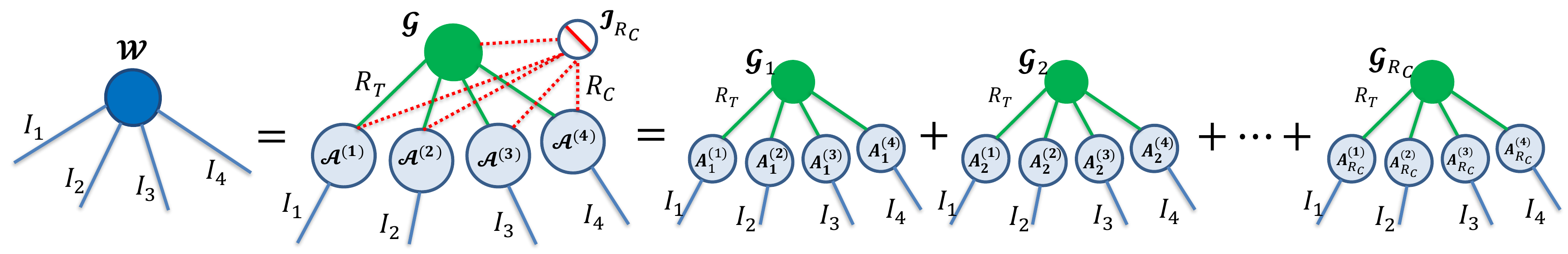}
 \vspace{-2ex}
  \caption{Diagrams for Block Term decomposition. It decomposes a tensor $\bm{\mathcal W}$ into a sum of several Tucker decompositions (in the right) with low Tucker-ranks ($R_T$). In the middle, there is a general representation by combining the identity tensor $\bm{\mathcal I}_{R_C}$.} \label{fig:m:BT_decomp}
 \end{figure*}
 
 \begin{figure*}[t]
\centering
\subfigure[Hierarchical block matrices to a tensor]{\label{fig:m:hbmat2tensor}
 \includegraphics[width=0.45\textwidth]{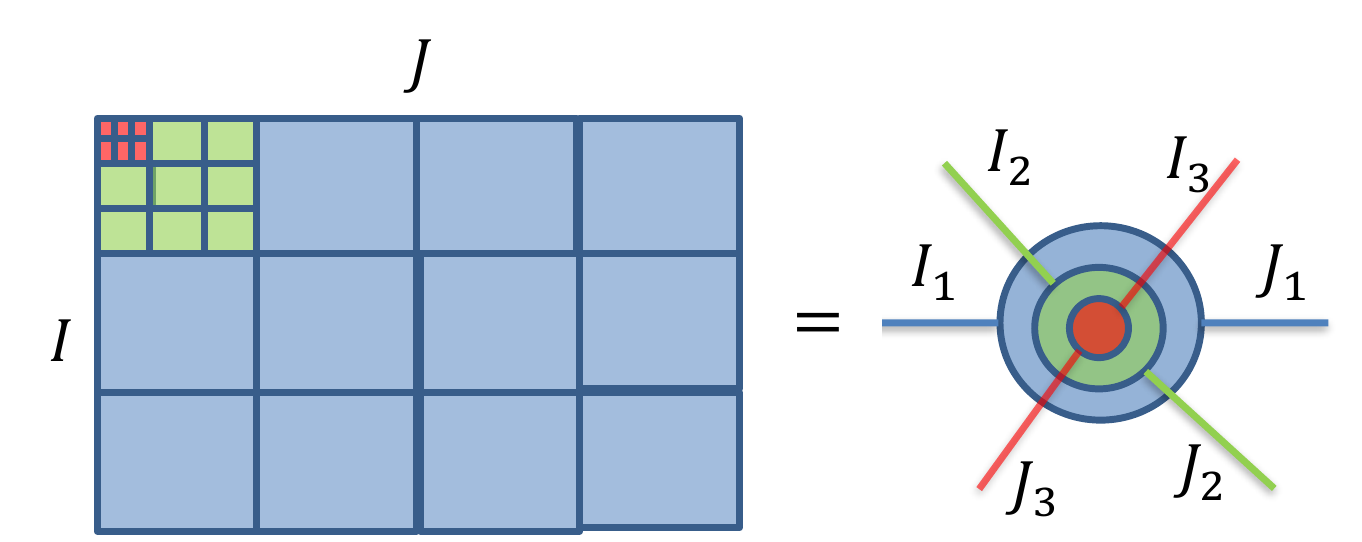}}
\subfigure[BT representation for a tensor]{\label{fig:m:tensor2BT}
 \includegraphics[width=0.53\textwidth]{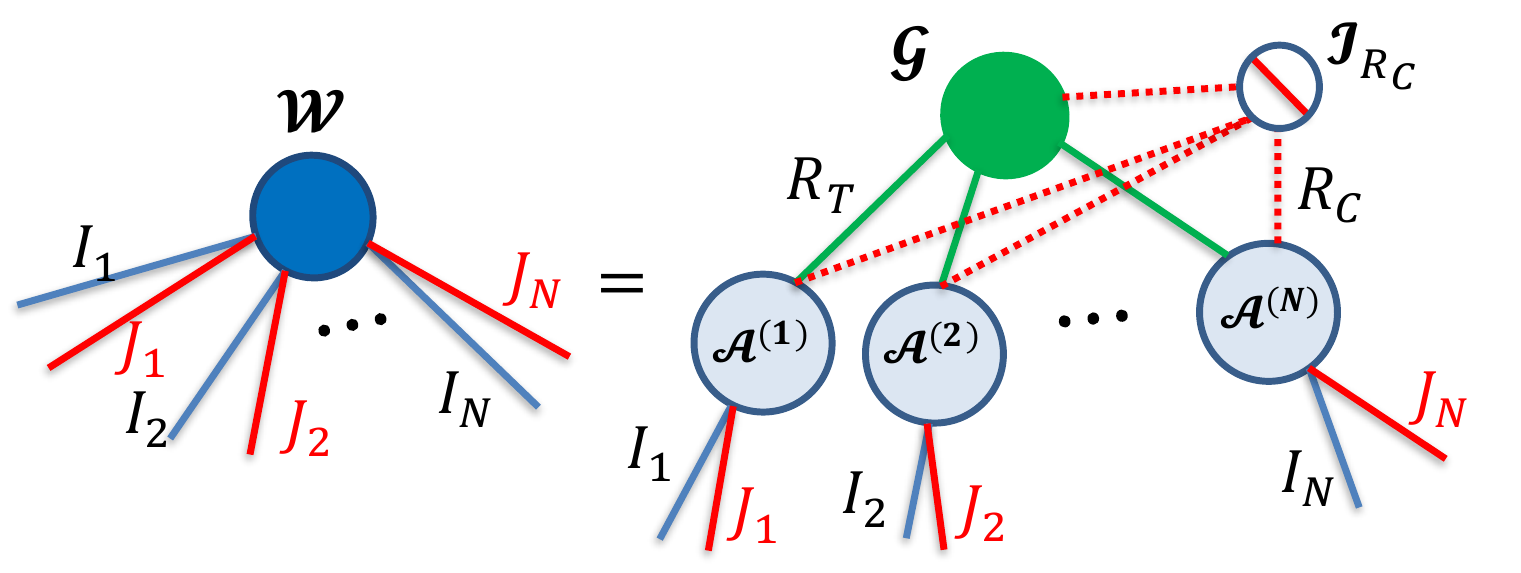}}
 \vspace{-2ex}
  \caption{Two steps of BT representation for a matrix. (a) First step: dividing the matrix into hierarchical block matrices and representing it as a higher-order tensor; (b) Second step: rearranging the tensor indices to keep the pairs of indices (coupled indices in each bock) together, e.g., $(I_1,J_1)$, $(I_2,J_2) \ldots$, then performing the BT decomposition.} \label{fig:m:BTrepresentation}
 \end{figure*}

There are usually two basic tensor decomposition methods: CP (CANDECOMP/PARAFAC) \cite{harshman1970foundations} and Tucker \cite{tucker1966some}. CP decomposes a tensor into a sum of several component rank-1 tensors; Tucker decomposes a tensor into a core tensor multiplied by a matrix along each mode. They are both classic yet their concentrations are different, as CP imposed a diagonal constraint on the core tensor of Tucker.
Thus, a more general decomposition called \textit{Block Term} (BT) decomposition, which combines CP and Tucker, has been proposed to make a trade-off between them \cite{de2008decompositions}.

The BT decomposition aims to decompose a tensor into a sum of several Tucker decompositions with low Tucker-ranks.
Specifically speaking, given an order-4 tensor $\bm{\mathcal W}\in\mathbb R^{I_1\times I_2\times I_3\times I_4}$, its BT decomposition can be represented by 6 nodes with special contractions, as illustrated in the middle of Fig.~\ref{fig:m:BT_decomp}. In the figure, $\bm{\mathcal I}_{R_C}$ denotes the order-5 dimension-$R_C$ identity tensor (only the super-diagonal positions have non-zero elements that are set to 1), $\bm{\mathcal G}\in\mathbb R^{ R_C\times R_T\times R_T\times R_T\times R_T}$ denotes the $R_C$ core tensors of Tucker decompositions and each $\bm{\mathcal A}^{(n)}\in\mathbb R^{R_C\times I_n\times R_T}$ denotes the $R_C$ corresponding factor matrices of Tucker decompositions. Moreover, each element of $\bm{\mathcal W}$ can be computed as follows:
\begin{align}
w_{i_1,i_2,i_3,i_4}=&\sum_{r_C=1}^{R_C}\sum_{r_{1},r_{2},r_{3},r_{4}=1}^{R_T,R_T,R_T,R_T} g_{r_C,r_{1},r_{2},r_{3},r_{4}} \nonumber\\ &\cdot a_{r_C,i_1,r_{1}}^{(1)} a_{r_C,i_2,r_{2}}^{(2)} a_{r_C,i_3,r_{3}}^{(3)} a_{r_C,i_4,r_{4}}^{(4)},
\end{align}
 where $R_T$ denotes the Tucker-rank (which means the Tucker-rank equals $\{R_T, R_T, R_T, R_T\}$) and $R_C$ represents the CP-rank. They are together called BT-ranks.

The advantages of BT decomposition mainly depend on the larger compression ratio on the number of elements in the original tensor and the compatibility with the benefits of both CP and Tucker. The reason is that when the Tucker-rank is equal to 1, the BT decomposition degenerates to CP decomposition; when the CP-rank equals 1, it degenerates to Tucker decomposition.
What's more, when compared with TT, it also has the attributes of commutativity and exponentiation, which will be introduced later.

\subsection{Block Term Representations for Matrices}

In this subsection, we introduce the BT representations for matrices, which means we use BT decomposition to represent the matrices that are ubiquitous in neural networks.
The procedure of doing BT representations for matrices mainly contains two steps: 

 \begin{itemize}
    \item Divide a matrix into hierarchical block matrices and represent it as a tensor \cite{cichocki2014era}. For example, in Fig.~\ref{fig:m:hbmat2tensor}, considering a matrix with size $I\times J$, we can divide it into $I_1\times J_1$ block matrices (in blue), then we subdivide each block matrix into $I_2\times J_2$ smaller block matrices (in red), and then we continue to subdivide each smaller block matrix (in green). We call a matrix with this structure the hierarchical block matrices, and now we view it as an order-6 tensor with dimension $I_1\times I_2\times I_3\times J_1\times J_2 \times J_3$, where $I=I_1 I_2  I_3$, $J=J_1 J_2 J_3$. This means the outermost matrix has size $I_1\times J_1$ and each of its elements is an $I_2\times J_2$ matrix whose every element is an another matrix with size  $I_3\times J_3$.
    \item Convert the higher-order tensor into a tensor with a new indices permutation that keeps the pairs of indices together, e.g., $(I_1,J_1)$, $(I_2,J_2) \ldots$,  then perform BT decomposition upon it. Since for each block matrix, the two dimensions (or indices) are coupled, we rearrange the indices of the tensor transformed from the original hierarchical block matrices to match the corresponding pairs of indices. The visualization can be seen intuitively in Fig.~\ref{fig:m:tensor2BT}. Then we carry out BT decomposition on it and as a result we can get the BT representation for the original matrix.
\end{itemize}

\section{BT-layer}

\begin{figure}[t]
\centering
\subfigure[FC-layer]{\label{fig:m:fclayer}
 \includegraphics[width=0.2\textwidth]{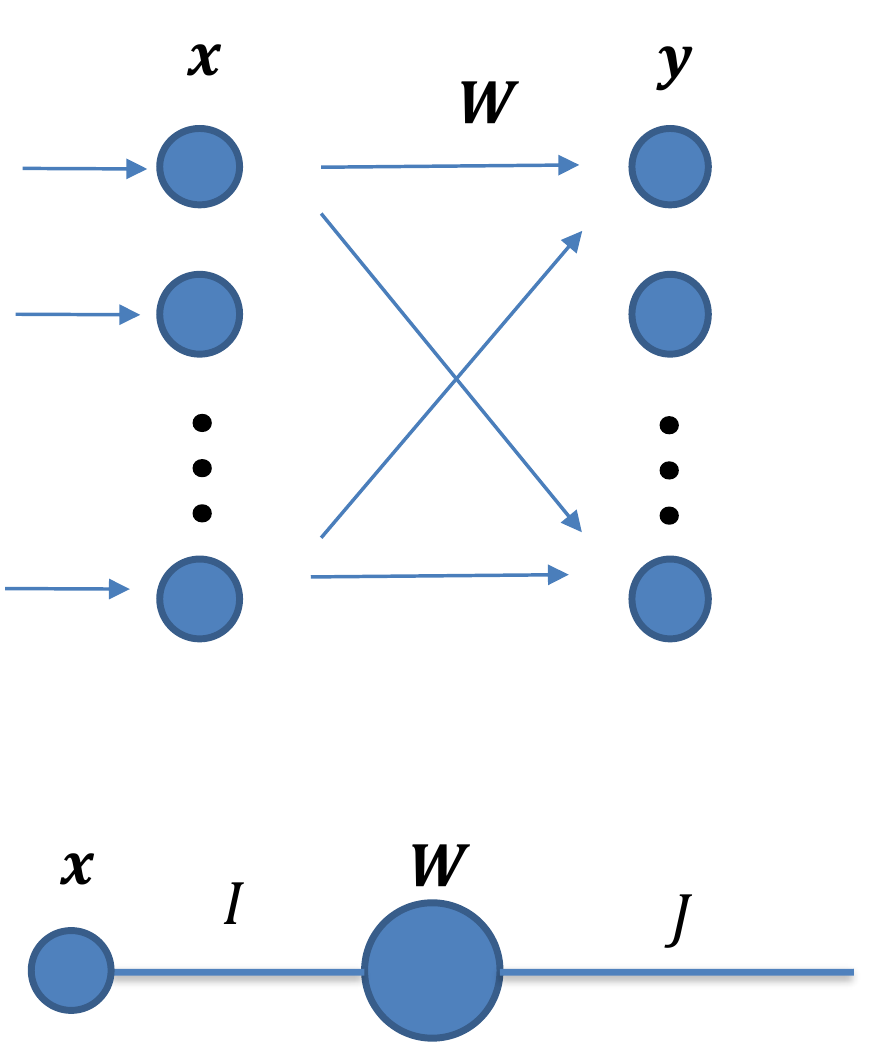}}
\subfigure[BT-layer]{\label{fig:m:BTlayer}
 \includegraphics[width=0.24\textwidth]{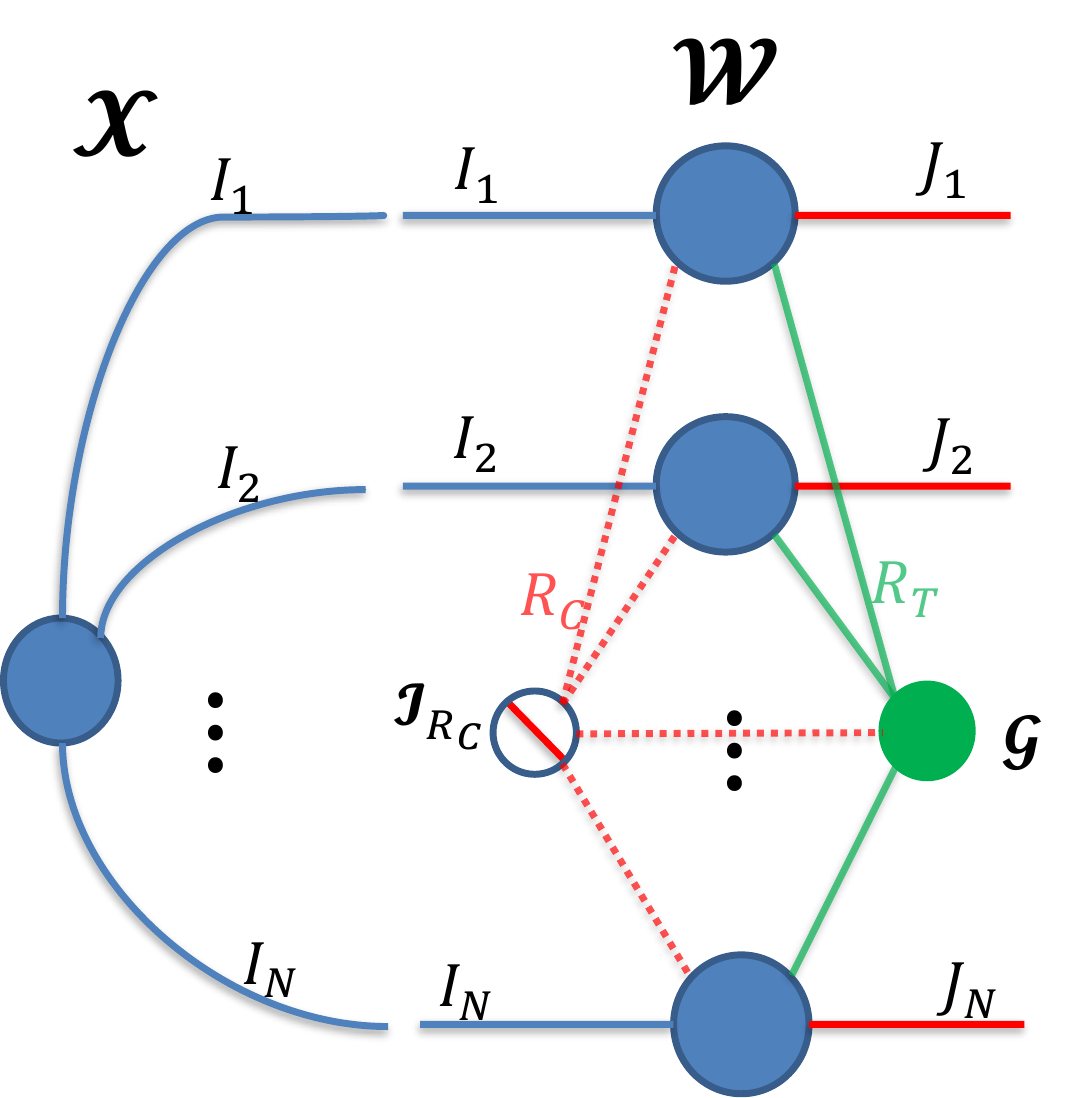}}
  \caption{Diagrams for FC-layer and BT-layer. (a) FC-layer in neural network case and tensor network case; (b) BT-layer with an order-$2N$ tensor weight.}
 \end{figure}

In this section we will introduce BT-layers of a neural network, where the weight matrices in fully-connected layers (FC-layers) are replaced with their BT representations.

The FC-layer is directly conducted by multiplying a weight matrix $\bm W\in\mathbb R^{J\times I}$ with an input vector $\bm x\in\mathbb R^{ I}$ and adding a bias vector $\bm b\in\mathbb R^{J}$:
\begin{align}
\bm{y}= \bm{W}\bm{x}+\bm{b},
\end{align}
where the output $\bm{y}$ is a vector with dimension-$J$. The diagrams for FC-layer is illustrated in Fig.~\ref{fig:m:fclayer}, where the above is the traditional neural network diagrams and the bottom is the tensor network diagrams. We can see that the nodes representing the data dimension and the edges representing the weight matrix in NN are changed into an edge with a number and a node in TN, respectively. The relationships among the nodes and the edges between NN and TN are dual. Since the addition operation is easy to solve, we omit the bias $\bm b$ here.

By contrast, the BT-layer, which is performed in tensor format, is conducted by multiplying an order-$2N$ tensor $\bm{\mathcal W} \in\mathbb R^{I_1\times J_1\times I_2\times J_2\times\ldots\times I_N\times J_N}$ and an order-$N$ tensor $\bm{ \mathcal X} \in\mathbb R^{I_1\times I_2\times\ldots\times I_N}$ and adding a bias order-$N$ tensor $\bm{\mathcal B}\in\mathbb R^{J_1\times J_2\times \ldots \times J_N}$:
\begin{align}
\bm{\mathcal Y}= \sum_{i_1,i_2,\ldots,i_N=1}^{I_1,I_2,\ldots,I_N}\bm{\mathcal W}_{i_1,\ast,i_2,\ast,\ldots,i_N,\ast} \mathcal X_{i_1,i_2,\ldots,i_N} +\bm{\mathcal B},
\end{align}
where $\bm{\mathcal Y}\in\mathbb R^{J_1\times J_2\times \ldots \times J_N}$ denotes the output tensor, $\bm{ \mathcal W}$ is the BT representation for the weight matrix and $\bm{ \mathcal X}$ represents the tensor format of the input vector which is equivalent to explicitly storing its elements by reshape operation. The diagrams for BT-layer is showed in Fig.~\ref{fig:m:BTlayer}, where the product of the contracted indices $\prod_{k=1}^NI_k=I$ denotes the input dimension and the product of the dangling edges $\prod_{k=1}^NJ_k=J$ denotes the output dimension. Here, we omit the bias, too.

In the learning step of BT-layer, we use the back-propagation procedure \cite{rumelhart1988learning}, the same as other layers in an Neural Network. However, please note that rather than learning the whole order-$2N$ tensor weight $\bm{\mathcal W}$ and performing BT representation upon it, we directly compute the core and factor matrices of $\bm{\mathcal W}$ instead. The core $\bm{\mathcal G}$ and factor matrices $\bm{\mathcal A}^{(n)}$ are illustrated in Fig.~\ref{fig:m:tensor2BT}.

\subsection{Complexity Analysis}
\paragraph{Number of parameters}
The total number of parameters of a BT-layer in Fig.~\ref{fig:m:BTlayer} is 
\begin{align}\label{m:eq:BT_paras}
    R_C(\sum_{k=1}^NI_kJ_kR_T+(R_T)^N).
\end{align}
By comparison, a FC-layer with the same input and output dimension has a total $\prod_{m=1}^NI_m\prod_{n=1}^NJ_n=IJ$ parameters.

\paragraph{Inference complexity}
Let's consider the situation in Fig.~\ref{fig:m:BTlayer}. This BT-layer has a inference complexity of $O(C_{BTL})$. We can write $C_{BTL}=R_C \cdot \tilde{C}_{BTL}$ where $\tilde{C}_{BTL}$ denotes the complexity of the tensor contractions between the input tensor $\bm{\mathcal X}$ and each block Tucker representation. To illustrate $\tilde{C}_{BTL}$ efficiently, we first compute the tensor contractions successively along indices $I_1, I_2,\ldots, I_N$, then compute the tensor contractions between the previous result tensor and the core tensor $\bm{\mathcal G}$. For the tensor contraction along index $I_k$, we have the following complexity:
\begin{align}
    \prod_{m=1}^{k-1}(J_mR_T)\prod_{n=k+1}^{N}I_n\cdot I_k\cdot J_kR_T.
\end{align}
Followed by the total complexity:
\begin{align}
    C_{BTL}=R_C \cdot\left(\sum_{k=1}^{N}\prod_{m=1}^{k}J_m\prod_{n=k}^{N}I_n(R_T)^{k}+J(R_T)^{N}\right).
\end{align}

If we assume $I_k=J_k$, then the inference complexity of a BT-layer can be simplified as 
$O(R_CNI(R_T)^{N}\max_k\{I_k\})$, $k=1, 2, \ldots, N$.

\paragraph{Complexity comparison}
Since the analysis process of training complexity is similar to it of inference complexity, here we omit it. We just report the final results in Table~\ref{model:table:complexity_comparison}, which also contains the results of FC-layer and TT-layer \cite{novikov2015tensorizing}.
Please note that $R_C, N, R_T, r$ and $m$ are significantly smaller than $I$ (or $J$). Therefore, it has a linear relationship with $I$ (or $J$), not the linear relationship with $IJ$.

\begin{table*}[t]\caption{Comparison of complexity and memory of an $I\times J$ FC-layer and its corresponding TT-layer and BT-layer, where $r$ denotes the TT-rank and $m = \max_k\{I_k, J_k\}$.}
\centering
\begin{tabular}{l|l|l}
  \Xhline{1.2pt}
  Operation & Time & Memory 
  \\\Xhline{1.2pt}
  FC forward 	& $O(IJ)$ & $O(IJ)$  \\
	FC backward & $O(IJ)$ & $O(IJ)$  \\
	TT forward 	& $O(N r^2 m\max\{I, J\})$ 		& $O(r \max\{I, J\})$ \\
	TT backward & $O(N^2 r^4 m\max\{I, J\})$ 		& $O(r^3 \max\{I, J\})$  \\
	BT forward 	& $O(R_CN(R_T)^{N}m\max\{I, J\})$  & $O((R_T)^N \max\{I, J\})$  \\
	BT backward & $O(R_CN^2(R_T)^{N}m\max\{I, J\})$ 	& $O((R_T)^N \max\{I, J\})$  \\
  \Xhline{1.2pt}
\end{tabular}\label{model:table:complexity_comparison}
\end{table*}

\section{Rank Bounds Analysis Between BT and TT}
It seems like that BT has more complicated structures or contractions than TT, so it should have more parameters than TT. Actually, because of its commutativity and exponential ranks, it can even have less parameters.

\begin{figure}[t]
  \centering
 \includegraphics[width=0.25\textwidth]{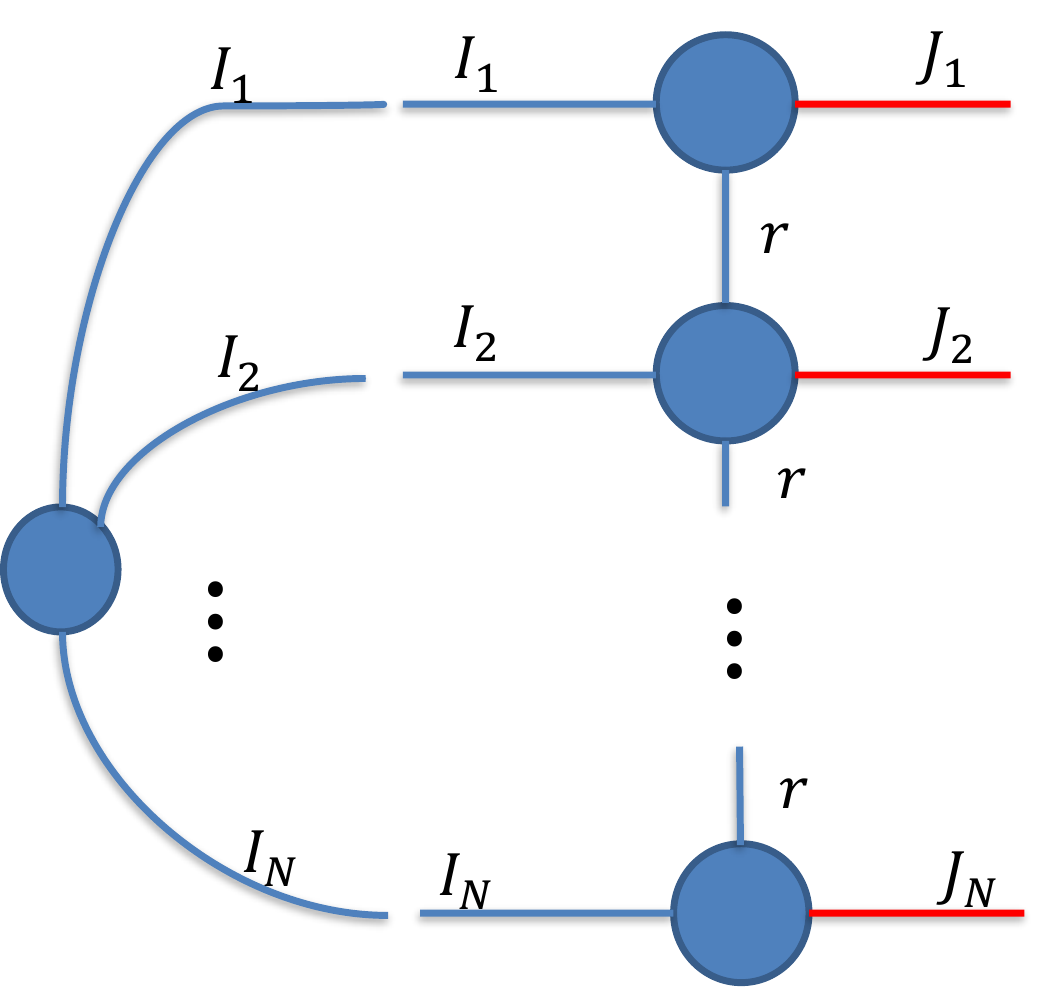}
 \vspace{-2ex}
 \caption{Diagrams for TT-layer.} \label{fig:m:TTlayer}
\end{figure}

\paragraph{Commutativity}
As for the TT decomposition, it is not commutative because different permutations of the input tensor contribute different TT-ranks of TT decomposition.
By comparison, BT decomposition is commutative because of its \textit{star form}  decomposition (Fig.~\ref{fig:m:BTlayer}), rather than \textit{linear form} in TT (Fig.~\ref{fig:m:TTlayer}). That means whatever permutations having been done, we will obtain a consistent and stable BT decomposition.

\paragraph{Exponential representation ability}
The connections between the first $k$ pairs of indices and the last $N-k$ of $\bm{\mathcal W}$ can be bounded by the rank of the following \textit{unfolding matrix} of $\bm{\mathcal W}$: $\bm W_{(\overline{i_1j_1,\ldots,i_kj_k};\overline{i_{k+1}j_{k+1},\ldots,i_Nj_N})}$. We simplify it as $\bm W_{[k]}$ for easy writing. Then we have
\begin{align}
    \text{rank}\ \bm W_{[k]}\leq R_C(R_T)^{\min\{k,N-k\}}.
\end{align}
If we suppose that $\bm W_{[k]}$ is sufficiently linear independent, then equal sign can be taken. This means we have stronger connections if we take more data modes into consideration. In other words, BT has exponential representation ability. This is also in conformity with our intuitive feeling.

By comparison, in the case of TT, the rank of  $\bm W_{[k]}$ satisfies
\begin{align}
    \text{rank}\ \bm W_{[k]}\leq r,
\end{align}
where $r$ denotes TT-rank. We can see that TT doesn't have exponential representation ability.





\section{Experiments}

In this section, we verify the effectiveness of our proposed BT-nets using various neural network architectures in terms of the compression ratio of BT-layers, accuracy and running time. Our experiments mainly contain the following three parts:
i) the performance of our series nets compared with the original nets and TT-nets on three datasets: MNIST \cite{lecun1998mnist}, Cifar10 \cite{krizhevsky2009learning}, ImageNet \cite{deng2012imagenet};
ii) the performance of our proposed BT-nets with respect to different $R_C$ and $R_T$ on Cifar10 dataset;
iii) the training time and inference time of our BT-layers and FC-layers.

In the results on the three datasets, we report the accuracy and the compression raito (Comp.R) of BT-layer with respect to different network architectures. We use ``$x$-BT$y$'' representing a BT-layer with CP-rank $(R_C)$ equaling $x$ and Tucker-rank $(R_T)$ equaling $y$. The Comp.R can be computed as follows:
\begin{align}
\text{Comp.R} = \frac{n_{original}}{n_{BT}},
\end{align}
where $n_{original}$ denotes the number of parameters in the replaced fully-connected layers of the original network and $n_{BT}$ denotes the number of parameters in the BT-layers of the BT-net.

\subsubsection{Implementation Details}
All network architectures used in the experiments can be implemented easily in TensorFlow and MxNet. The experiments are performed on a NVIDIA TITAN Xp GPU and a Tesla k40m GPU.
We train all networks from scratch with stochastic gradient descent with momentum of 0.9. To avoid the influence of random initialization and the problem of gradient vanishing or exploding, it's necessary to add a batch normalization layer after BT-layer.

\subsection{Results on MNIST}
The MNIST dataset of handwritten digits is composed of 70,000 gray  $28\times 28$ images in 10 classes. There are 60,000 training examples and 10,000 test examples.
As a baseline we use the convolutional neural network LeNet-5
\cite{lecun1998gradient}, with two convolutional, activation function (tanh) and max-pooling layers followed by two fully-connected layers (FC-layers) of size $800\times 500$ and $500\times 10$.

\subsubsection{Network Architectures} 
We fix the convolutional part of the original network and just replace the first FC-layer with BT-layer. 
The BT-layer reshapes the input and output tensors as $5\times 5\times 8\times 4$ and $5\times 5\times 5\times 4$ tensors respectively. As the prediction task for MNIST dataset is quite easy, we simply use one block term decomposition (e.g., Tucker decomposition) with Tucker-rank equaling to 2 and 3 for a large compression ratio. We also run the TT-net as a comparison by replacing the first FC-layer with TT-layer, where the output dimension is substituted as $5\times 5\times 5\times 4$ and the TT-rank is set as 2.

Table~\ref{e:table:acc_on_mnist} reports the results on MNIST dataset. The first column represents the different network architectures, the middle two columns represent the number of parameters in the first FC-layer (or its alternatives) and the compression ratio respectively. The last column represents the accuracy on test dataset. We can see at first glance that the number of parameters in the FC-layer can be reduced from 800$\times$500 to 228 in ``1-BT2'' and the compression ratio can up to 1754, with only 0.03\% decrease in accuracy. The compression ratio of the entire network is 14.01.
We can also observe that ``1-BT3'', with 399 parameters in BT-layer, has the same accuracy as the baseline while TT-net lost 0.03\% in performance on about the same order of magnitude of the parameter amount.


\begin{table}[t]\caption{Results on MNIST}
\centering
\begin{tabular}{c|ccc}
  \Xhline{1.2pt}
  Architecture & \# paras & Comp.R & Acc (\%)  \\ \Xhline{1.2pt}
  baseline & 800$\times$500 & 1 & 99.17 \\
  TT2 & 342 & 1169&  99.14 \\
  1-BT2 & \textbf{228} & \textbf{1754}  & 99.14 \\
  1-BT3 & 399& 1002  & \textbf{99.18} \\
  \Xhline{1.2pt}
\end{tabular}\label{e:table:acc_on_mnist}
\end{table}








\subsection{Results on Cifar10}




The Cifar10 dataset consists of 60,000 32$\times$32 color images in 10 classes, such as airplane, bird, cat, etc, with 6,000 images per class. There are 50,000 training images and 10,000 test images.
We refer to the tensorflow model\footnote{https://github.com/tensorflow/models/tree/master/tutorials\\/image/cifar10} as baseline, which consists of two convolutional, local respond normarlization (lrn) and max-pooling layers followed by three FC-layers of size $2304\times 384$, $384\times 192$ and $192\times 10$.

\subsubsection{Network Architectures}
We similarly replace the first FC-layer with BT-layer which treats the input and output dimensions as $6\times 6\times 8\times 8$ and $6\times 4\times 4\times 4$ respectively. TT-net is also just replacing the first FC-layer with TT-layer which has the same output dimension reshaping as BT-layer.
We let the CP-rank vary from 1 to 8 and the Tucker-rank vary from 1 to 3 in the BT-layer and let TT-rank equals 2 and 8 in TT-layer.

Some results of Cifar10 dataset are reported in Table~\ref{e:table:acc_on_cifar10}  (others can be found in Fig.~\ref{e:fig:BTrank}). We can see that when using ``1-BT2'' structure, the compression ratio is up to 3351 at the cost of about 1\% reduction in accuracy. By comparison, the compression ratio of ``TT2'' is only 2457 with almost the same accuracy as BT-layer. In response to the increase in the architecture's complexity, we observe that ``4-BT3'' has a larger compression ratio while preserving a better accuracy at the same time compared with ``TT8''. And ``4-BT3'' can has a total compression  ratio of 2.98.


\begin{table}[t]\caption{Results on Cifar10}
\centering
\begin{tabular}{c|ccc}
  \Xhline{1.2pt}
  Architecture & \# paras& Comp.R & Acc (\%)  \\ \Xhline{1.2pt}
  Baseline & 2304$\times$384  & 1  & \textbf{85.99} \\
  TT2 & 360  & 2457   & 84.90 \\
  TT8 &  4128 & 214 &  85.70 \\
  1-BT2 & \textbf{264} & \textbf{3351} & 84.95 \\
  4-BT2 & 1056  & 838 & 85.47 \\
  4-BT3 & 1812  & 488 & 85.83 \\ 
  \Xhline{1.2pt}
\end{tabular}\label{e:table:acc_on_cifar10}
\end{table}

\subsection{Results on ImageNet}

The ILSVRC 2012 (ImageNet) is a large dataset which consists of 1.2 million images for training and 50,000 for validation and comprises 1000 classes. As a baseline, we use the AlexNet architecture\footnote{https://github.com/apache/incubator-mxnet/tree/master\\/example/image-classification}, which has three FC-layers of size $6400\times 4096$, $4096\times 4096$ and $4096\times 1000$.

\subsubsection{Network Architectures}
We replace the first FC-layer with BT-layer where the input and output dimensions are reshaped as $10\times 10\times 8\times 8$ and $8\times 8\times 8\times 8$ respectively. The same dimension  reshaping is performed in TT-layer as well. As in the Cifar10 case, we experiment with two groups of variations (simple and complex) of the BT-layer and the TT-layer. Accordingly, we choose ``1-BT2'' and ``4-BT2'' as BT-layers and set TT-rank as 2 and 8 in TT-layers.

In Table~\ref{e:table:acc_on_ImageNet} we report the compression ratio, Top-1 and Top-5 accuracy on different architectures. From the results we see that BT-layer in the best case (``4-BT2'') can get a  compression ratio of 11070 (from 6400$\times$4096 parameters to 2368) on amount of parameters while achieving a slightly better Top-1 and Top-5 accuracy than baseline at the same time. The total compression ratio of the entire network is 2.06. By comparison, ``TT8'' only gets a  compression ratio of 2528 and even about 1\% accuracy drop. Similarly, ``1-BT2'' gets a compression factor of more than 40,000 with 2.2\% decrease in Top-5 accuracy, better than ``TT2''.
Please note that all the experiments on the Imagenet are performed without fine-tuning.


\begin{table}[t]\caption{Results on ImageNet}
\centering
\begin{tabular}{c|ccc}
   \Xhline{1.2pt}
  Archi- & Comp.R & Acc (\%) & Acc (\%)  \\
        tecture &  & Top-1 & Top-5  
  \\\Xhline{1.2pt}
  Baseline & 1 & 56.17 & 79.62 \\
  TT2 & 30340 & 52.14 & 76.40 \\
  TT8 & 2528 & 55.11  & 78.61  \\
  1-BT2 & \textbf{44281} & 53.20 & 77.38 \\
  4-BT2 & 11070 & \textbf{56.48} & \textbf{79.69} \\ 
  \Xhline{1.2pt}
\end{tabular}\label{e:table:acc_on_ImageNet}
\end{table}
\subsection{Flexibility of BT-ranks}

\begin{figure}[t]
  \centering
 \includegraphics[width=0.4\textwidth]{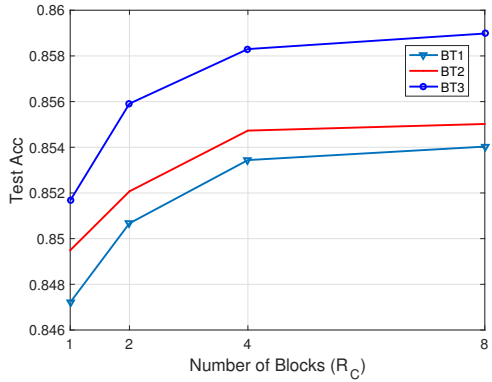}
 \vspace{-2ex}
 \caption{The test accuracy versus the BT-ranks on Cifar10 dataset.} \label{e:fig:BTrank}
\end{figure}

As BT-layer has two kinds of ranks, e.g., CP-rank $(R_C)$ and Tucker-rank $(R_T)$, we regard the Cifar10 dataset as example to study their impacts on the performance.
The network architectures are designed as previously stated: the CP-rank varies from 1 to 8 and Tucker-rank 1 to 3.

The results are reported in Fig.~\ref{e:fig:BTrank}. We can intuitively see that the larger BT-ranks, the higher test accuracy. In details we observe that when the number of blocks (CP-rank) is small, the accuracy curves rise quickly, but when the CP-rank becomes large, the curves is almost horizontal. The condition in Tucker-rank is analogous. However, if the Tucker-rank is too large, the BT-layer will become quite complex and its number of parameter will increase sharply because the amount parameters of the core are exponential (see Formula~\ref{m:eq:BT_paras}). Similarly, if the CP-rank is too large, there is no significantly increase in accuracy.
Thus, if we want to get a better performance, we need to consider both and let them be appropriate values.

\subsection{Results of Running Time}



\begin{figure}[t]
\centering
\subfigure[Training time]{\label{e:fig:train_time}
 \includegraphics[width=0.46\textwidth]{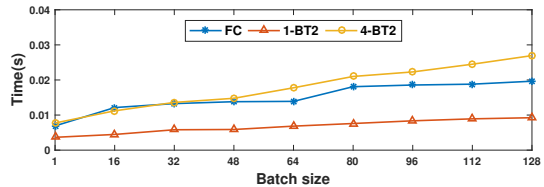}}
\subfigure[Inference time]{\label{e:fig:inference_time}
 \includegraphics[width=0.46\textwidth]{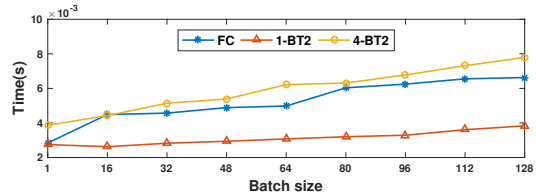}}
 \vspace{-2ex}
  \caption{Running time of $6400\times 4096$ FC-layer and its corresponding BT-layers.}
  \label{e:fig:running_time}
 \end{figure}

Since BT-layer can reduce the number of parameters in FC-layer largely, how does it have the time consumption? In order to explore this question, we test on a single $6400\times 4096$ FC-layer and its corresponding BT-layers which have the same architectures as in the ImageNet experiment.
These benchmark experiments are performed on TensorFlow on a single Tesla k40m GPU.

The results of the training time and inference time with different batchsize are showed in Fig.~\ref{e:fig:running_time}.
 From the results we can intuitively observe that when BT-ranks are small, BT-layer has a significant acceleration effect compared with FC-layer. When BT-ranks are increased to let the BT-net catching the original performance, the time cost in BT-layer is almost the same as in FC-layers.

\section{Conclusions and Future Works}


We have proposed the BT-nets, a new network architecture in which the commonly used fully-connected layers are replaced with the BT-layers. In the BT-layers, the large weight matrices in the FC-layers are represented as tensors whose indices are rearranged, and block term decompositions are performed on these tensors.
Since BT decomposition has exponential ranks, BT-layers are able to preserve the representation power of FC-layers as much as possible.
Our experiments demonstrate that we can obtain a very large compression ratio of the number of parameters of a single layer while almost maintaining the performance.

As for future work, we plan to compress other parts of DNNs to achieve the purpose of compressing the entire network. And we will consider combining the BT-nets with other compression techniques, such as pruning and binarization.

\section{Acknowledgement}
This paper was in part supported by Grants from the Natural Science Foundation of China (No. 61572111), the National High Technology Research and Development Program of China (863 Program) (No. 2015AA015408), 
a 985 Project of UESTC (No. A1098531023601041) and a Fundamental Research Funds for the Central Universities of China (No. A03017023701).

\bibliographystyle{aaai}
\bibliography{BTNets}
\end{document}